# Power Transformer Fault Prediction Based on Knowledge Graphs

Chao Wang[1], Zhuo Chen[1], Ziyan Zhang[2], Chiyi Li[1], Kai Song[2]

[1]State Grid Chongqing Electric Power Company
[2]School of Information Science and Engineering, Chongqing Jiaotong University

**Abstract**

In this paper, we address the challenge of learning with limited fault data for power transformers. Traditional operation and maintenance tools lack effective predictive capabilities for potential faults. The scarcity of extensive fault data makes it difficult to apply machine learning techniques effectively. To solve this problem, we propose a novel approach that leverages the knowledge graph (KG) technology in combination with gradient boosting decision trees (GBDT). This method is designed to efficiently learn from a small set of high-dimensional data, integrating various factors influencing transformer faults and historical operational data. Our approach enables accurate safe state assessments and fault analyses of power transformers despite the limited fault characteristic data. Experimental results demonstrate that this method outperforms other learning approaches in prediction accuracy, such as artificial neural networks (ANN) and logistic regression (LR). Furthermore, it offers significant improvements in progressiveness, practicality, and potential for widespread application.

## 1 Introduction

Power transformer is the main equipment in the transmission link of power grid, and it is also the key object of power grid operation and inspection business. In order to ensure its operational safety, the State Grid Corporation has issued various related maintenance management regulations and standard processes, such as the State Grid Corporation Transformer Maintenance Management Regulations, etc. Statistically analyzed, transformer failures have been reported in the past few years. Statistically analyzed, transformer failures are mainly due to poor material and technology [1-4], human factors, poor operating environment [5], and excessive electrical loads [6-10], and common failure phenomena include not limited to: transformer oil temperature exceeds the threshold. Common fault phenomena include, but are not limited to: transformer oil temperature exceeding the threshold value, abnormal transformer oil level, abnormal transformer odor and appearance, transformer overload causing over-voltage or over-load, transformer bushing aging or failure, sharp and harsh abnormal operation sound, bath current and voltage increase caused by poor core grounding, and oil leakage caused by transformer material or process problems, etc. [1]. At present, with the introduction of advanced technologies such as 5G and IoT, online real-time monitoring of transformer temperature, noise, voltage and current, local discharge and other operating conditions has basically been realized by installing oil chromatography monitoring sensors, core grounding current sensors, bushing monitoring sensors, temperature sensors, acoustic fingerprinting monitoring sensors, etc. on power transformers to address the above mentioned common faults [12 -16], and online real-time monitoring of transformer temperature, noise, voltage, current, local discharge and other operating conditions is also possible for transformers [13]. 16], and for transformer operation and maintenance, also on-line informationization expert system, basically realized the transformer equipment account management, operation status monitoring and fault statistical analysis and other business applications. But for the hidden faults are still basically in the middle of the alarm or after the statistics stage, the lack of fault analysis and prediction of information technology means [17].

In recent years, machine learning models, such as support vector machines (SVMs) and artificial neural networks (ANNs), have been popularly used as factor-driven methods for routine fault analysis and prediction, and have shown a certain degree of competitiveness in knowledge learning tasks [18]. Compared with simple statistical models, machine learning is more flexible to high-dimensional training data and is able to capture more feature information from historical transformer data sets [19]. This is conducive to obtaining better results for analyzing transformer operating conditions and stability. However, training an ideal machine learning model requires a large amount of valid sample data, which poses a challenge in scenarios where only a small amount of fault data is available for the transformer. Knowledge graph (KG) technology has been widely used in various intelligent reasoning scenarios and machine learning tasks because of its triple prediction capability by constructing semantic networks of world entities [20]. Gradient boosted decision tree (GBDT) can utilize the feature crossover of the factors in the historical data set to provide more useful feature information for training the knowledge graph, and can learn from a small set of high-dimensional data, while taking into account the influence of different component failure factors and the correlation between all historical data [21].

Therefore, in order to realize the fault analysis and prediction of power transformer when limited fault characteristic data (a small amount of high-dimensional historical data) are available, this paper puts forward a method to build the risk prediction of transformer operation state based on knowledge graph and gradient lifting decision tree technology, and realizes the safety state evaluation and fault analysis and prediction of power transformer by theoretical analysis and experimental research, and verifies the effectiveness and accuracy of the proposed method to predict the safety risk of transformer operation through the real historical data set of transformer actual operation.

## 2 Knowledge Graph and its transformer fault prediction advantages

### 2.1 Knowledge Graph technology

Knowledge graph is a semantic network of real-world entities and describes the relationships between them. Various heterogeneous and complex datasets can be integrated into such a knowledge structure to explicitly capture the relationships between entities and their attributes. In principle, a knowledge graph is represented as a directed graph containing nodes and directed edges, where a node represents an entity, denoted as $E$, and an edge connecting a pair of nodes represents a relation, denoted as $R$. By referring to real-world knowledge as a set of knowledge, a knowledge graph is essentially a collection of knowledge. Each of these knowledge can be represented by a triple $(h, r, t)$, where $h$ and $t$ are head and tail entities $(h, t \in E)$, respectively, and $r$ is the relation between $h$ and $t$ $(r \in R)$. Thus, the triple $(h, r, t)$ denotes that there exists a knowledge of the relation $r$ from the head entity $h$ to the tail entity $t$ [22].

Knowledge graphs have been applied to various machine learning tasks such as knowledge quizzing, logical reasoning [23-25], recomm-enddder systems and information retrieval. With the rapid development of deep learning and embedded technology, it has become common to use representation learning techniques, such as low-dimensional vectors, to identify the entities and relationships in knowledge graphs. Using knowledge graph techniques, various effective data analysis operations (e.g., inference and prediction) can be performed to deal with complex practical learning tasks.

Many advanced knowledge graph applications have been proposed and developed at home and abroad, including Freebase, Yago, Nell and WordNet. However, due to the limitation of the incompleteness of the existing technology products, these knowledge graph techniques cannot be effectively applied to complex industrial scenarios. Recently, many researches are working on predicting missing triples by modeling existing knowledge to mitigate the effect of incompleteness. Among these results, "translation-based" modeling has become a relatively popular approach due to its good results in terms of accuracy and scalability. This scalability is important for complex prediction tasks in knowledge graphs. Therefore, knowledge graph is potentially a better approach to tackle the fault risk prediction class of tasks because of its triple prediction capability.

## 2.2 Advantages of Knowledge Graph for Transformer Fault Prediction

The performance of most prediction models such as ANNs and SVMs depends on the size and quality of the sample data. In the study scenario, the real data of transformer accidents or failures are small, and the data may be noisy and heterogeneous. For small datasets, if not properly trained, the model (e.g., neural network) may under- or over-fit resulting in poor prediction performance [26]. In short, learning models cannot use small training datasets to explore complex relationships between factors. For traditional statistical-based models, such as logistic regression and Bayesian networks, their performance deteriorates as the slope factor increases, and these models are ineffective for handling high-dimensional datasets [27].

Knowledge graph is a potential solution to the above problem, as its training process relies on a collection of triples derived from existing data, each of which is a training sample of the knowledge graph, defined as $(h, r, t)$, it contains the relationship $r$ between the head entity $h$ and the tail entity $t$, and constructs the semantic network of knowledge graph of transformer equipment by using the translation-based model. Based on the existing research results, this paper realizes the "translation" transformation from the head entity $h$ to the tail entity $t$ by training a model with minimum loss and adopting the relation $r$. In principle, a small group of equipment failure or accident records can generate enough triplets for training, and each entity in the triplet can be defined as the feature intersection (that is, high-dimensional vector) of equipment steady state data. The research method in this paper not only pays attention to the similarity of data level, but also emphasizes the correlation between historical fault or accident data, so it is more suitable for dealing with small data sets and high-dimensional data structures.

The knowledge graph is a specific structural model that can learn entity characteristics and relationships between entities, where each historical operating condition assessment value record is represented as a specific entity. In addition, the representation vectors of the entities in the knowledge graph can be crosscut with the features in the historical operational state data. These features can be enriched by utilizing the correlation between historical data. In this paper, we utilize the advantages of combining knowledge graph and equipment safety and stability assessment, and the basic idea is relatively simple, i.e., under the framework of knowledge graph, a set of ternary groups are generated from equipment fault or accident records for training, and a semantic network of equipment operating state is constructed for inference and prediction [28].

## 3 Small-sample based knowledge graph complementation model

### 3.1 Meta-learning method

Meta-learning is the study of systematic observation of multiple learning tasks and learning new tasks from them, with the goal of generalizing the distribution of tasks and optimizing batches of tasks. The goal is to generalize the distribution of tasks to optimize batches of tasks. Each task has a corresponding learning problem, and well-performing tasks can improve the learning efficiency and generalize to a small sample of problems without overfitting. As shown in Table 1, there are three types of datasets in meta-learning, namely, the support set $S$, the query set $Q$ and auxiliary set $A$, And according to the N-way and K-shot methods, each set is sampled to form a training set $T$.

The relationship is a common part of the support set and the query set. The purpose of the experiment is to migrate the relationship information shared by the support set and the query set to the triple of the missing tail entity, and realize the completion of small sample knowledge by migrating the relationship-specific meta-information, so that the model can learn important knowledge quickly by using the relationship element and gradient meta-information. Among them, the relational element represents the relationship between the support set and the query set that connects the head entity and the tail entity; The gradient element is a gradient that supports the centralized relational element. By changing the relational element through the gradient element, the minimum loss is achieved and the learning process is accelerated. In addition, on the one hand, the meta-information of specific relationship transfers the common information from triplet to incomplete triplet; On the other hand, by observing a small number of examples, the learning process within the task is accelerated.

The meta-learning model advocates cross-task learning and adaptation to new tasks, aiming to learn task unknown models on relational tasks, not only for containing task-specific models. The model handles the small sample learning problem through meta-training and testing phases by first defining the existing complete entity-relationship triad as the meta-training set $D_{tm}$ for the complementary task, and then the knowledge graph triad to be complemented is defined as the meta-test set $D_{tst}$ for the complementation task, Finally, all triplets in $D_{tm}$ and $D_{tst}$ are initialized to get triplets $(h, r, t) \in G$ represented by vectors. $G$ is the set of ternary samples, and $h, t \in E$, $E$ is the set of entities, r $r \in R$, $R$ is the set of relations.

In the case of $D_{tr}$ in which the triples with the same relation are grouped into the same set, then the set is the relational task corresponding to the relation $T_r, T_r \in T, T \sim p(T)$, $p(T)$ is the set of tasks consisting of all relational tasks. A task is randomly selected from the task set $p(T)$, and $N$ triple samples are taken as the support set $S_r$ of the task, and the remaining samples are taken as the query set $Q_r$ of the task, and the number of samples in the support set is smaller than that in the query set. The specific algorithm of meta-learning model is as follows:

---

Input: Relational task training set $T_{tm}$.

Output: Parameters of the embedded layer emb, parameters $\varphi$ of the meta-learner, new relation element $R'$.

When the training set $T_{tm}$ is empty.

In the training set $T_{tm}$ define a task in $T_r = \{S_r, Q_r\}$

From the support set $S_r$ take the relational element in $R$.

Compute the current task $S_r$ the loss function and score function.

From the relational element $R$ obtaining the gradient element in $G$.

According to the gradient element $G$ updating the relational element $R$.

Compute the current task query set $Q_r$ the loss function of the

According to the task query set $Q_r$ the loss function updates the parameters of the learner $\varphi$ and the embedded layer parameter emb.

End the learning process.

Output the new relational element $R'$.

By means of $L$ layer fully connected neural network to extract entity specific relational elements:

$$x^0 = h_i \oplus t_i$$
$$x^1 = \sigma(W^l x^{l-1} + b^l)$$
$$R_{(h_i,t_i)} = W^L x^{l-1} + b^l$$

Among them, $h_i \in R^d$ is the embedding of the head entity $h_i$ dimension $d$, $t_i \in R^d$ is the $t_i$ dimension of the tail entity is the embedding of $d$, and $L$ is the number of layers of the neural network $l \in \{1, \cdots, L-1\}$, $W^l 1 b^l$ is the weights and biases of $l$-layer neurons, and the activation $\sigma$ is calculated by LeakyReLU, where $x \oplus y$ represents the splicing of vectors $x$ and $y$, and $R_{(h_i,t_i)}$ relational elements representing specific entities $h_i$ and $t_i$.

For multiple entity pairs $(h_i, t_i)$ of a specific relation element, the final relation element is generated by the specific relation elements of all entity pairs in the current task:

$$R_{T_r} = \frac{\sum_{i1}^{K} R_{(h_i,t_i)}}{K}$$

where, in the case of $K$ denotes the number of relational elements.

### 3.2 Evaluation method

In order to evaluate the updating effect of gradient tuples on relation tuples, it is necessary to construct a score function to evaluate the effective ranking of entity pairs under a specific relation, and to compute the loss function for the current task. For this purpose, the core idea of the embedded method is applied to the embedded learner to confirm the validity of the true ranking of ternaries in the knowledge graph.

In the task $\tau_r$, calculate the score of each entity in support set $S_r$ for $(h_i, t_i)$:

$$s_{(h_i,t_i)} = \|h_i + R_{T_r} - t_i\|$$

where, the $\| x \|$ denotes the vector $x$ of the L2 norm. Assuming that the triplet $(h, r, t)$ composed of the head entity, the relation and the tail entity satisfies $h + r = t$, the distance between $h + r$ and $t$ can be compared by using the score function to minimize it. This idea can be applied to the small sample link prediction task, but because there is no direct relationship embedding in the task, the relationship embedding $r$ is replaced by the relationship element $R_{\tau_r} \tau$.

For each ternary score function, the following loss function is set:

$$L(S_r) = \sum_{(h_i,t_i \in S_r)} \left[ \gamma + S_{(h_i,t_i)} - S_{(h_i,t_i')} \right]_+$$

Where $[x]_+$ represents a positive triple of $x$, $\gamma$ represents an edge superparameter, $S_{(h_i,t_i')}$ is the score of the entity pair $(h_i, t_i) \in S_r$ in the current support set, and $S(h_i, t_i')$ is the score of the negative sample in the support set, both of which satisfy $(h_i, r, t_i') \notin G$.

Since the task $T_r$ denotes that the model is able to encode the correct ternary correctly, and $L(S_r)$ ) is small, so it will be based on $L(S_r)$ of $R(T_r)$ the gradient is regarded as the gradient element $G(T_r)$:

$$G(T_r) = \nabla_{R_r} L(S_r)$$

According to the gradient update rule, the relation element is updated quickly:

$$R(T_r)' = R(T_r) - \beta \nabla_{R_r} L(S_R)$$

where, the $\beta$ denotes the step size of the gradient element during the relational element operation.

When the query set is scored by the code learner, after the updated relational element $R(T_r)'$ is obtained, it is transferred to the sample of the query set $Q_r = \{(h_j, t_j)\}$, and the score and loss in the query set are calculated. Similarly, operate on the support set:

$$S_{(h_j,t_j)} = \|h_j + R'_{\tau_r} - t_j\|$$

$$L(Q_r) = \sum_{(h_j t_j)} \left[ \gamma + S_{(h_j,t_j)} - S_{(h_j,t_j')} \right]_+$$

where $L(Q_r)$ is the training objective for minimization, and its loss can be used to update the model.

The training objective is to minimize the loss function, and the sum of the losses for all tasks in a small batch of samples can be expressed as

$$\mathbf{L} = \sum_{(S_r, Q_r)} \mathbf{L}(\boldsymbol{Q_r})$$

## 4 Knowledge Graph Implementation of Transformer Fault Prediction Methods

### 4.1 Translation-based Knowledge Graph Training with Few Samples

Knowledge graph is regarded as a semantic network paradigm for feature discovery tasks, reflecting various relationships between entities in the network. "Translation-based" modeling is a popular knowledge representation learning approach for building knowledge graphs. The basic idea is to encode entities and their relationships into a continuous low-dimensional vector space [29]. Given a new entity, the trained knowledge graph model can be used to determine the relationship between the newly added entity and the existing entities. In this paper, the idea of the sample less training process is applied to the construction of the knowledge graph, which is combined with the historical record of equipment operation. Sample less learning is a kind of meta-learning that learns how to distinguish different classes from a dataset that consists of many classes (each with several samples) [30]. In its training process, samples from different classes are used as support data and are paired with samples as query data. This sample less learning model predicts whether the pairs of samples belong to the same class, thus maximizing the limited number of samples in each class to train the model [31].

In the scenario of this paper, the transformer equipment represented by entities can be divided into two classes: normal equipment and faulty equipment, and it is assumed that the relationship between the same class and different classes is "similar" and "non-similar", specifically: the relationship between the faulty and normal equipment is "non-similar", and the relationship between two devices in faulty state or two devices in normal state is "similar". The knowledge graph can be used as a classifier in the sample less learning task model, and gives the result of whether the pair of operational state values belongs to the same class or not. By combining the knowledge graph and the history of operational state evaluation values through the above sample less training idea, the prediction of entity relationships can be realized.

Each equipment operating state in the historical data set can be represented as a node in the constructed knowledge graph. The semantic network of the transformer is constructed as follows. Any equipment operating state entity belongs to a certain category, so any pair of entities is associated with a specific relationship, and labeled according to the actual situation, when all the entities involved are labeled, the semantic network construction is complete, any equipment operating state (i.e., entity) can be associated with all the other equipment operating states, according to their categories through entity relationships. Based on Fig. 1, it is not necessary to find out all the existing relationships between any pair of equipment operating states, but only a part of the existing relationships can be used to form a semantic network that can be reasoned and predicted [33].

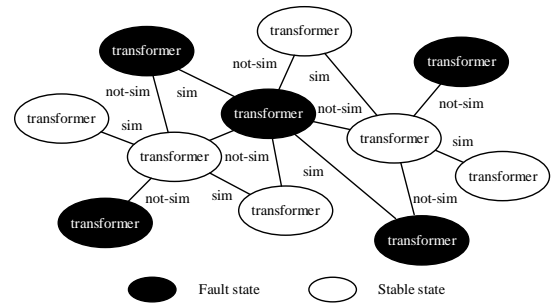

**Fig. 1 Semantic network of operation state of transformer equipment**

Each semantic vector of stable operation state of transformer equipment is represented by a historical data, which takes into account both the correlation information between historical data and the influence of equipment failure factors. The constructed semantic network can be used to determine the relationship between a new equipment stability condition and each condition in the historical data. In this way, it is possible to predict the operating state of a new transformer.

### 4.2 Data set construction

The dataset consists of a triad, which is the smallest unit describing the entities and relationships in the knowledge graph, including the head entity, relationship and tail entity, usually denoted as (head, relationship, tail). The model is trained by constructing pairs of transformer devices according to the small-sample learning training process. In this work, the knowledge graph that serves as a classifier gives predictions in the form of relationships between pairs of entities. Thus, the triad consists of the categories of the equipment's operational stability to determine the relationship between them.

One transformer device is randomly taken as the head entity and another transformer device is taken as the tail entity, which are combined into a triad, and "similar" or "non-similar" relationships are added according to the actual equipment operation state classes. Based on Fig. 2, any two different equipment operation stable states from the same class can form a ternary group with "similar" relationship. Any pair of equipment states from two classes can form a triad with a "different" relationship. Randomly selecting the equipment operation steady states to form different triples, until the number of triples reaches the upper limit of combinations.

Accordingly, the number of triples constructed based on the existing operating history of the device will exceed the number of existing operating history records of the device. Even if the number of history records is small, there are still enough triples to train the knowledge graph model. This avoids the problem of having many small training samples. In addition, when there is an imbalance between the two classes of records, existing learning-based models usually do not learn well the class with fewer records. Using the dataset construction process described above, the problem of imbalanced dataset can be mitigated by giving more chances for the fewer records to appear.

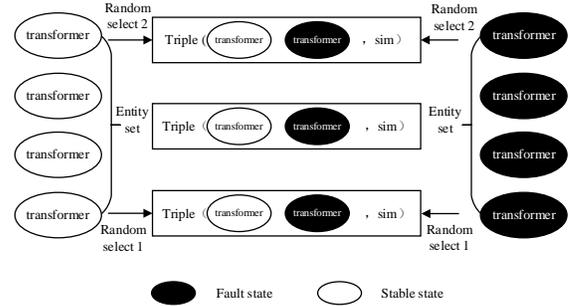

**Fig. 2 Construction process of transformer triples based on the training idea of few shot learning**

### 4.3 Feature Intersection

In a general knowledge graph, entities are represented by semantic vectors. In the scenario of this paper, a set of equipment accident or failure related factor values in the historical records can be used for feature crossover to distinguish or characterize the transformer state. Gradient boosting decision trees can effectively represent various entities in the knowledge graph by constructing such feature intersections [34].

Gradient Boosting Decision Trees use the principles of gradient descent and boosting to explore feature intersection solutions by creating a set of decision trees. Each new tree is constructed based on the residency between the actual results and the predictions given by the previously constructed old trees [35]. Given the history of equipment operating states in faulty or steady state, the gradient boosting decision tree is trained to capture the feature crossover capability by creating a series of regression trees, which are constructed as follows:

$L(y_i, f(x_i))$ is the loss function of the decision tree; $m$ is the number of decision trees; $r_{mi}$ is the residual of the $i$-th sample in the $m$-th decision tree in the construction process, and its formula is defined as

$$r_{mi} = -\left[\frac{\partial L(y_i, f(x_i))}{\partial f(x_i)}\right] f(x) = f_{m-1}(x)$$

A tree is trained starting with a weak classifier and using the gradient as the residual. $C_{mj}$ is the best value that fits the $j$-th leaf node, and $R_{mj}$ is the set of all regions of the leaf node, which is given by

$$c_{mj} = \arg\min_c \sum_{x_i \in R_{mj}} L(y_i, f_{m-1}(x_i) + c)$$

In order to obtain a smaller error and a stronger classifier, the origin part is added to the $m$-th new regression tree, eq.

$$f_m(x) = f_{m-1}(x) + \sum_{j=1}^{J} c_{mj} I(x \in R_{mj})$$

The whole model is updated by stacking all the decision trees together with the equation

$$\hat{f}(x) = f_M(x) = \sum^{M} \sum^{J} c_{mj} I(x \in R_{mj})$$

Then, the trained gradient boosting decision tree model is used to construct a feature crossover over the transformer operation status history records. Taking each record as an input sample, the gradient boosting decision tree model explores its feature intersection by tracking the sample in the regression tree. In the gradient boosting decision tree, each transformer operating condition sample can be tracked in each tree structure according to its different factor values. As each tree node splits for feature selection, the multilayer structure of nodes naturally constitutes an effective feature crossover. In other words, samples can reach different tree nodes in each tree according to their original feature values, and these tree trajectories are effective feature intersection strategies. In this way, different transformer operating states are effectively distinguished.

### 4.4 Knowledge Graph Prediction Model

The "translation-based" knowledge graph is based on the assumption that the header entity vectors plus the relationship vectors are equal to the tail entity vectors. Setting the vector $h$ and vector $t$ to represent the head entity and the tail entity, and the vector $r$ to represent the relationship between the pair of entities, can be expressed by equation (5)

$$h + r = t$$

After the above feature intersection process, the vector $S_i = \{s_{i1}, s_{i2}, s_{i3}, s_{i4}, \ldots, s_{in}\}$ represents a transformer device with $n$ dimensions, and each dimension of the vector can represent a feature of the operating state of the transformer, and in order to show that different features have different impacts on the operating state of the transformer, Multiply an $n$-dimenSional weighting vector $W = \{w_1, w_2, w_3, w_4, \ldots, w_n\}$ by the transformer vector $S_i$, in this knowledge graph model, the head entity and the tail entity are denoted by

$$h = W \cdot S_i$$
$$t = W \cdot S_j$$

According to the assumption of ideal conditions, it can be derived that

$$|h + r - t| = 0$$

Equation (8) is the ideal case, in the actual case, $|h + r - t|$ is not zero, we need to determine the relationship between a pair of entities through the ratio, the minimum value of the relationship is the result of this model. In order to train this knowledge graph model, the triples containing the wrong relations are combined in order to calculate the loss value during the model training. As mentioned before, the relationship between transformers is {similar, non-similar}, for the real relationship between entities $h$ and

$t$, let it be $r_t$; The unreal relationship between entities $h$ and $t$ is denoted as $r_f$; If the real relationship between two transformers is similar, $r_t$ means that the vectors are similar, and $r_f$ means that the vectors are not similar; On the contrary, if the real relationship between two transformers is non-similar, $r_t$ means non-similar, and $r_f$ means similar. According to the above assumption, $r_t$ and $r_f$ are calculated in equation $|h + r^-t|$ with reference to equations (8) and (9).

$$e_1 = |h + r_t - t|$$

$$e_2 = |h + r_f - t|$$

After calculating the two values, the model takes the relationship with the smaller gap as the prediction. If $e_1$ is greater than $e_2$, it means that the gap between the entity pair and the true relationship is greater than the gap between the entity pair and the non-true relationship, which must be less than the gap between the non-true relationship in the real situation [3]. So in this case, we need to consider the loss, which is calculated as follows

$$loss = e_1 - e_2$$

On the contrary, if $e_1$ is smaller than $e_2$, it means that the distance between the entity pair and the real relationship is smaller than the distance between the entity pair and the non-real relationship. This situation is consistent with the facts, the loss can be counted as zero.

Summarizing the above two cases, the loss function of the model can be calculated as

$$loss = \max(|h + r_t - t| - |h + r_f - t|, 0)$$

The architecture of the prediction model for the knowledge graph in the paper is shown in Fig. 3.

Its losers are entity vector triples obtained from the gradient boosting decision tree model. The gradient boosting decision tree is treated with nonlinear relationships and gives a reasonable crossover of different features. The knowledge graph model designed in this paper takes into account the historical records and the influence of different feature factors, and its entities are adequately represented.

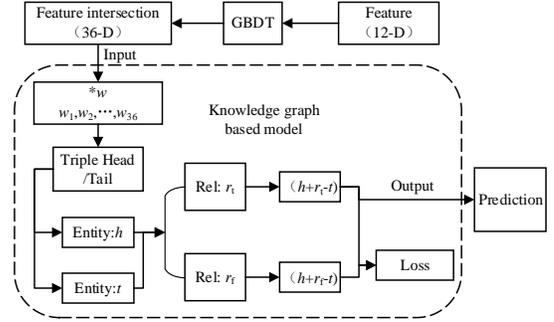

**Fig. 3 Basic structure of GBDTKG prediction model**

### 4.5 Failure rate

Through the above process, a well-trained gradient boosting decision tree + knowledge graph prediction model is obtained, which can be used to carry out fault analysis and prediction through the entity relationship between a transformer and all the transformers in the history records, and the prediction results can be expressed by the transformer failure rate (TFR).

Considering that the same entity has different meanings when it is the tail and the head in the triple, this paper constructs two entity pairs, including a newly put transformer and a transformer in operation. Because the relationship between entities in the knowledge graph is directed, the new transformer can be marked as the head entity or the tail entity. For each new transformer, record its number according to its matching method: if the new transformer is similar to the historical fault transformer, record it as $L_s$; If the new transformer is not similar to the historical fault transformer, it is recorded as $L_d$; If the new transformer is similar to the historical steady-state transformer, it is marked as $S_s$; If the new transformer is not similar to the historical

steady-state transformer, it is marked as $S_d$. Transformer TFR calculation formula is

$$TFR = (L_s + S_d)/(2NS)$$

In Eq. (13), NS is the number of historical transformers.

## 5 Experimental results and analysis

### 5.1 Historical data collection of transformer status

Firstly, the historical operation data of transformers are collected through historical files or databases, and 131 records are collected for each transformer in stable state and fault state. Each transformer record includes the following attributes. Each transformer record includes the following attributes: load current, oil temperature, oil level, gas, oil color, sound normal value, appearance cleanliness, silicone discoloration and so on.

Next, the 262 records were divided into two parts. The 262 records are divided into two parts: 10 records in steady state and 10 records in fault state for final effect testing, and 121 records in steady state and 121 records in fault state for model training.

Then, 6000 triples were generated, including 3000 "similar" and 3000 "non-similar" triples, and divided into training and testing sets according to the ratio of 7:3, and the number of "similar" and "non-similar" triples in the training and testing sets were the same.

After completing the above data preparation work, the transformer entity in the training set can be used to train the gradient boosting decision tree model, and then test the prediction model with the test set, and the transformers in the test set need to ensure that they are not in the historical data record.

### 5.2 Experimental results and analysis of prediction accuracy

Firstly, a random gradient optimization tool-Adam optimizer is used to optimize the test parameters, and the learning rate is set to 0.2. The gradient lifting decision tree model is trained. According to the stability of transformer, the model learns how to combine different influencing factors in a reasonable way. The learning rate is set to 0.001, 0.005 and 0.01 respectively, and the knowledge graph prediction model in this paper is tested. The test results include loss value and accuracy rate.

The test results show that the model will collect the roses in each case, when the learning rate is 0.001, the experimental accuracy is around 84%, which indicates that the prediction model does not fall into the local optimum when the learning rate is 0.001 on the data set, and the training results are reliable. The highest accuracy of predicting the relationship during the training process of this model is more than 85%, and the average accuracy reaches 84.71%, which indicates that the accuracy of this model in predicting the relationship between two transformer entities is 84.71%.

Validation experiments based on ANN and LR were conducted on the same ternary dataset to compare the effects of different models, and the results show that the prediction accuracy of the present knowledge graph prediction model is very high. The results are with the highest training accuracy of 64.86% for LR model and 62.47% for ANN model. The experimental results show that, for the scenarios in the paper, the model accuracy curve of the

In order to verify the effect of gradient boosting decision tree on the quality of prediction models, extended experiments are conducted, in which the entities are directly represented by historical data, and the gradient boosting decision tree is not used for feature crossing. The training results are shown in Fig. 6, and the model is closed with an accuracy of 74.98%. This result shows that the use of gradient boosting decision tree is indeed effective in improving the ability of predicting relationships and leads to the full utilization of historical data. On the other hand, even without the help of gradient boosting decision trees, the knowledge graph-only prediction model still performs better than the ANN and LR models. The accuracy results obtained by different models on the test set are shown in Table 1. The test results show that the proposed Gradient Boosting Decision Tree + Knowledge Graph prediction model can integrate the advantages of the Gradient Boosting Decision Tree as a feature engineering method for the structure of the Knowledge Graph, and the overall results are better than other models.

**Table 1 Test results on relation prediction**

| Model | Accuracy /% |
| --- | --- |
| Artificial neural network model | 62.47 |
| Logistic regression model | 64.86 |
| Knowledge Graph Model | 74.98 |
| GBDT + knowledge graph + Spectral model | 84.71 |

### 5.3 Experimental results and analysis of predicted fault occurrence rate

As mentioned above, 20 historical records are used as the test data set, where 20 test records represent 20 test transformer entities and 242 training records represent 242 training transformer entities. The goal of the test is to predict the failure rate of a test transformer using a trained Gradient Boosting Decision Tree + Knowledge Graph prediction model, which is designed to determine the relationship between the test transformer and all training transformers. As a result of the model's risk prediction, the TFR of each test transformer can be calculated by Eq. (13), and the results of the model's tests on 20 predicted records are shown in Table 2.

Among the test results, there are 10 records of transformer in faulty state (sample number 1~10) and 10 records of transformer in stable state (sample number 11~20). Test transformers with TFR greater than 0.5 are treated as faulty states, and the model predicts the faulty state of the transformer.

Both faulty and stable states are accurately judged. If we choose the slippage slope with TFR greater than 0.85, the model correctly predicts 9 fault samples, and the TFR of the stable transformer samples is around 0.1, which indicates that the model has better ability to predict the stable and fault states of the transformer.

**Table 2 Prediction results of samples**

| No. | FSS (Ls) | FSN (Ld) | SSS (ss) | SSN (sd) | TFR |
| --- | --- | --- | --- | --- | --- |
| 1 | 82 | 160 | 2 | 240 | 0.6652 |
| 2 | 180 | 62 | 2 | 240 | 0.8677 |
| 3 | 184 | 58 | 10 | 232 | 0.8595 |
| 4 | 197 | 45 | 14 | 228 | 0.8780 |
| 5 | 179 | 63 | 1 | 241 | 0.8677 |
| 6 | 177 | 65 | 1 | 241 | 0.8636 |
| 7 | 189 | 45 | 14 | 228 | 0.8780 |
| 8 | 180 | 65 | 1 | 241 | 0.8636 |
| 9 | 29 | 53 | 9 | 233 | 0.8719 |
| 10 | 27 | 62 | 3 | 239 | 0.8657 |
| 11 | 65 | 213 | 221 | 21 | 0.1033 |
| 12 | 65 | 215 | 232 | 10 | 0.0764 |
| 13 | 49 | 177 | 222 | 20 | 0.1756 |
| 14 | 25 | 177 | 222 | 20 | 0.1756 |
| 15 | 26 | 193 | 228 | 14 | 0.1301 |
| 16 | 25 | 217 | 230 | 12 | 0.0764 |
| 17 | 26 | 216 | 228 | 14 | 0.0826 |
| 18 | 25 | 217 | 230 | 12 | 0.0764 |
| 19 | 25 | 217 | 229 | 13 | 0.0785 |
| 20 | 26 | 216 | 232 | 10 | 0.0743 |

## 6 Conclusion

In this paper, a new transformer fault risk prediction method based on knowledge graph is proposed, which uses knowledge graph to construct a transformer equipment operation state risk prediction model, and adopts gradient boosting decision tree to construct feature intersection to improve the accuracy of relationship prediction. The specific conclusions are as follows.

1) A power transformer fault prediction model based on gradient boosting decision tree and knowledge graph is proposed. Based on knowledge graph technology, a semantic network model for transformer failure risk prediction is constructed by utilizing a small amount of fault data of transformer equipment and the relationship and correlation between different factors; the application of gradient boosting decision tree reveals the deeper influence of different factors through the intersection of many features affecting the safe operation of the transformer, and the introduction of the gradient boosting decision tree to search for the nonlinear relationship between factors can provide more information for the knowledge graph to improve the accuracy of the relationship prediction. The introduction of the gradient boosting decision tree to find the nonlinear relationship between the factors can provide more entity information for the knowledge graph.

2) The use of Knowledge Graph + Gradient Boosting Decision Tree technique is more suitable for dealing with feature-rich samples, rather than relying on orthogonal features that require a large number of samples. In this way, reliable transformer operation fault prediction can be performed with a small amount of historical data rather than a large number of samples. This is an important advantage of the proposed method over other existing methods, because most of the actual transformer operating environments cannot provide a large number of fault state data records.

3) The proposed method not only takes into account the influencing factors related to the occurrence of faults, but also considers the correlation information between historical data. Therefore, the task of transformer fault risk prediction can benefit from the experience of all historical data, and thus reliable fault risk prediction results can be obtained.

4) The experimental results on the real historical data of transformer operation show that the model is highly effective and accurate in predicting the safety risk of transformer operation on a small amount of high-dimensional historical data. Therefore, the proposed knowledge graph-based transformer failure risk prediction method has good practicality and significance.